\documentclass[conference]{IEEEtran}
\usepackage{cite}
\usepackage{amsmath,amssymb,amsfonts}
\usepackage{algorithmic}
\usepackage{graphicx}
\usepackage{pgfplots}
\usepackage{stfloats}
\usepackage{subfig}
\usepackage{textcomp}
\usepackage{xcolor}
\def\BibTeX{{\rm B\kern-.05em{\sc i\kern-.025em b}\kern-.08em
    T\kern-.1667em\lower.7ex\hbox{E}\kern-.125emX}}

\usepackage{hyperref}
\usepackage[T1]{fontenc} 
\usepackage[utf8]{inputenc}

\usepgfplotslibrary{groupplots}
\usetikzlibrary{matrix,positioning}
\pgfplotsset{compat=newest}

\newlength{\size}
\begin{document}

\title{Evolving Evaluation Functions \\for Collectible Card Game AI
}

\author{
\IEEEauthorblockN{Rados{\l}aw Miernik}
\IEEEauthorblockA{\textit{Institute of Computer Science} \\
\textit{University of Wroc{\l}aw}\\
Wroc{\l}aw, Poland \\
radekmie@gmail.com}
\and
\IEEEauthorblockN{Jakub Kowalski}
\IEEEauthorblockA{\textit{Institute of Computer Science} \\
\textit{University of Wroc{\l}aw}\\
Wroc{\l}aw, Poland \\
jko@cs.uni.wroc.pl}
}

\maketitle

%

\begin{abstract}

In this work, we presented a study regarding two important aspects of evolving feature-based game evaluation functions: the choice of genome representation and the choice of opponent used to test the model.

We compared three representations. One simpler and more limited, based on a vector of weights that are used in a linear combination of predefined game features. And two more complex, based on binary and n-ary trees.

On top of this test, we also investigated the influence of fitness defined as a simulation-based function that: plays against a fixed weak opponent, plays against a fixed strong opponent, and plays against the best individual from the previous population.
 
For a testbed, we have chosen a recently popular domain of digital collectible card games. We encoded our experiments in a programming game, Legends of Code and Magic, used in Strategy Card Game AI Competition. However, as the problems stated are of general nature we are convinced that our observations are applicable in the other domains as well.
\end{abstract}

\begin{IEEEkeywords}
Evolutionary Algorithms, Evaluation Functions, Collectible Card Games, Genetic Programming, Strategy Card Game AI Competition, Legends of Code and Magic
\end{IEEEkeywords}

\section{Introduction}

The vast majority of game playing algorithms require some form of game state evaluation -- usually in the form of a heuristic function that estimates the value or goodness of a position in a game tree.
This applies not only to classic min-max based methods, as used for early Chess champions \cite{Campbell2002Deep}, but also in modern approaches, based on deep neural networks combined with reinforcement learning \cite{silver2018general}. 
The idea of estimating the quality of a game state using a linear combination of human-defined features was proven effective a long time ago \cite{samuel1959some}, and is still popular, even when applied to modern video games \cite{garcia2020optimizing,justesen2016online}.

One of the recently popular game-related AI testbeds are Collectible Card Games (e.g., \emph{Hearthstone} \cite{Blizzard2004Hearthstone}, \emph{The Elder Scrolls: Legends} \cite{Bethesda2017TESL}). Because they combine imperfect information, randomness, long-term planning, and massive action space, they impose many interesting challenges \cite{hoover2019many}. 
Also, the specific form of game state (combining global board features with features of individual cards) makes such games particularly suited for various forms of feature-based state evaluation utilizing human player knowledge.

The motivation for our research was to compare two models for evolving game state evaluation functions: a simpler and more limited, based on a vector of weights that are used in a linear combination of predefined game features; a more complex and non-linear, based on a tree representation with feature values in leaves and mathematical operations in nodes.

As a testbed, we have chosen a domain of digital collectible card games. We encoded our models in a programming game, Legends of Code and Magic, used in various competitions played on the CodinGame.com platform and IEEE CEC and IEEE COG conferences.

During the initial experiments, we observed some interesting behaviors of both representations regarding ``forgetting'' previously learned knowledge. As these observations were related to the chosen goal of evolution, we performed additional tests comparing three fitness functions: playing against a fixed weak opponent, playing against a fixed strong opponent, and playing against the best individual from the previous population.

This paper is structured as follows. 
In the next section, we present the related work describing usages of evolution for game state evaluation and the domain of AI for collectible card games.
In Section III, we provide the details of our model describing the representations and fitness functions.
The following two sections present our experiments and discuss the results in the above-mentioned topics: comparing tree versus vector representations and fixed opponent versus progressive fitness calculation.
In the last section, we conclude our research and give perspectives for future work.

\section{Background}

\subsection{Evolving evaluation functions}

Evolutionary algorithms are used for game playing in two main contexts.
One, based on the Rolling Horizon algorithm (RHEA) \cite{perez2013rolling}, utilizes evolution as an open-loop game tree search algorithm.
The other, a more classic approach, employs evolution offline, to learn parameters of some model, usually a game state evaluation function.
Such function can be further used as a heuristic in alpha-beta, RHEA, some variants of MCTS \cite{Browne2012ASurvey}, or other algorithms, and its quality directly translates to the agent's power.

\newpage

The common approach is to define a list of game state features and evolve a vector of associated weights, so that they closely approximate the probability of winning the game from the given state.
This type of parameter-learning evolution has been applied to numerous games, including Chess \cite{david2013genetic}, Checkers \cite{kusiak2007evolutionary}, TORCS \cite{salem2018evolving}, Hearthstone \cite{santos2017monte}.

Although treating parameters as vectors and evolving their values using genetic algorithms is more straightforward, some research uses tree structures and genetic programming instead for this purpose.

The majority of genetic programming applications in games are an evolution of standalone agents.
It was successfully applied in various board games, e.g., Chess \cite{gross2002evolving, hauptman2005endchess} and Reversi \cite{benbassat2012evolving}.

It is also possible to combine genetic programming with other algorithms and techniques.
One example is to combine it with neural networks to evolve Checkers agents \cite{khan2008developing}.
Another is to evolve the evaluation function alone and combine it with an existing algorithm, e.g., MCTS, instead of evolving fully-featured agents.
Such an evolution was successfully applied in games of varying complexity, e.g., Checkers \cite{benbassat2011evolving} and Chess \cite{ferrer1995using}.

Another aspect is measuring the quality of an evaluation function.
Because it has to compare the strength of the agents, usually a simulation-based approach is used, combined with a large number of repetitions to ensure the stability of obtained results.
Thus, such an evolution scheme is very computationally expensive.

\subsection{AI for collectible card games}

\emph{Collectible Card Game} (CCG) is a broad genre of both board and digital games. Although the mechanics differ between games,  basic rules are usually similar. First, two players with their \emph{decks} draw an initial set of cards into their \emph{hands}. Then, the main game starts in a turn-based manner. A single \emph{turn} consists of a few \emph{actions}, like playing a card from the hand or using an onboard card. The game ends as soon as one of the players wins, most often by getting his opponent's health to zero.

Recently the domain has become popular as an AI testbed, resulting in a number of competitions \cite{HearthstoneAICompetition,janusz2017helping,LOCMPage}, and publications focusing on agent development, deckbuilding, and game balancing.

Usually, agents are based on the Monte Carlo Tree Search algorithm \cite{Browne2012ASurvey} (as it is known to perform well in noisy environments with imperfect information), combined with some form of state evaluation either based on expert knowledge and heuristics as in \cite{santos2017monte}, or neural networks \cite{zhang2017improving}. 
An interesting approach combining MCTS with supervised learning of neural networks to learn the game state representation based on the word embeddings \cite{mikolov2013distributed} of the actual card descriptions is described in \cite{swiechowski2018improving}.

The deckbuilding task for the \emph{constructed} game mode (in which players can prepare their decks offline, selecting cards from a wide range of possibilities) has been tackled in several works. The most common approach is to use evolution (e.g., \cite{bjorke2017deckbuilding} for Magic: The Gathering, \cite{garcia2016evolutionary,bhatt2018exploring} for Hearthstone), combined with testing against a small number of predefined human-made opponent decks. 
Alternatively, a neural network-based approach for Hearthstone has been presented in \cite{chen2018q}. 

Balancing, in the context of collectible card games, usually means slight modifications of card statistics (e.g., attack, health, cost) to prevent overpowered decks. 
MAP-Elites with Sliding Boundaries algorithm has been proposed for this task \cite{Fontaine2019Mapping}.
The study in \cite{deSilva2019evolving} proposes multi-objective evolution, trying to minimize the magnitude of changes.

Other attempts focus on understanding the game while it is played. One can try to predict cards that the opponent is likely to play during a game \cite{bursztein2016legend}, or predict a probability of winning as in \cite{grad2017helping}.

\subsection{Legends of Code and Magic}

Legends of Code and Magic (LoCM) \cite{LOCMPage} is a small CCG, designed especially to perform AI research, as it is much simpler to handle by the agents, and thus allows testing more sophisticated algorithms and quickly implement theoretical ideas.
The game contains 160 cards, and all cards' effects are deterministic, thus the nondeterminism is introduced only via the ordering of cards and unknown opponent's deck.
The game is played in the fair arena mode, i.e., before every game, both players create their 30-card decks decks secretly from the symmetrical yet limited card choices (so-called \emph{draft} phase). 
After the draft phase, the main part of the game (called the \emph{battle} phase) begins, in which the cards are played according to the rules, and the goal is to defeat the opponent.
LoCM is also used in \emph{Strategy Card Game AI Competition} co-organized with CEC and COG conferences since 2019.

As for today, there is not much research for the \emph{arena} game mode.
The problem of deckbuilding has been approached in \cite{Kowalski2020EvolutionaryApproach} using an active genes evolutionary algorithm that in each generation learns only a specific part of its genome.
An approach using deep reinforcement learning is presented in \cite{vieira2020drafting}.  
Usually, the deckbuilding phase is solved via predefined card ordering or some heuristic evaluation of the card's strength. As a result, each draft turn from given three cards to choose, the strongest is taken.
The winner of the 2020 COG competition has been described in \cite{Witkowski20LOCM}. The authors used static card weights computed using harmony search for the draft phase and MCTS with various enhancements, including an opponent prediction for the battle game phase.
The authors of \cite{montoliu2020efficient} perform study on application of Online Evolution Planning approach \cite{justesen2016online} using heuristic tuned via N-Tuple Bandit Evolutionary algorithm \cite{lucas2018n}. 
Other playing approaches include minimax, best-first-search, MCTS, and rule-based decision making.

\section{Evaluating Card Games}

\subsection{Representation}\label{sec:repr_details}

We have developed three distinct representations: \texttt{Linear}, \texttt{BinaryTree}, and \texttt{Tree}.
Each one implemented two operations: \texttt{evalCard} (used for the draft phase and as a part of the state evaluation) and \texttt{evalState} (used for the battle phase).

To limit the vast space of possible evaluation functions, the final state evaluation is a sum of \texttt{evalState} (based on features related with the global board state) and \texttt{evalCard} (depending on card-related features) for each own card on the board, minus \texttt{evalCard} for each opponent card.
It is a common simplification, used in e.g., \cite{santos2017monte}. 

\begin{itemize}
\item \texttt{Linear}, is a constant-size vector of doubles.
Each gene (from 1 to 20) encodes a weight of the corresponding feature.
The first 12 are game state features (used for \texttt{evalState}), 6 for each of two players: current mana, deck size, health, max mana, number of cards to draw next turn, and next rune (an indicator for an additional draw as in \cite{Bethesda2017TESL}).
The other 8 are card features (used for \texttt{evalCard}): attack, defense, and a flag for each of the keywords, encoded as 1.0 when set and 0.0 when not.
The final evaluation is a sum of features multiplied by their corresponding weights.
\item \texttt{BinaryTree}, is a pair of binary trees, encoding state and card evaluation respectively.
The leaf nodes are either constants (singular double) or features, same as in \texttt{Linear}.
Both trees have the same set of binary operators (nodes): addition ($l+r$, where $l$ and $r$ are the values of left and right subtree respectively), multiplication ($l*r$), subtraction ($l-r$), maximum ($max(l, r)$), and minimum ($min(l, r)$).
The final state evaluation is calculated recursively accumulating the tree.
\item \texttt{Tree}, is a pair of n-ary trees, encoding state and card evaluation respectively.
The leaf nodes are identical to the ones in \texttt{BinaryTree}.
Operators are no longer binary, but n-ary instead -- each operator stores a vector of subtrees.
Available operators are addition ($\sum$), multiplication ($\prod$), maximum ($max$), and minimum ($min$).
Additionally, to ensure that every operation stays well defined, all subtree vectors are guaranteed to be nonempty.
To make subtraction possible, there is one additional, unary operator: negation ($-x$, where $x$ is the value of its subtree).
The final evaluation is calculated recursively accumulating the tree.
\end{itemize}

Note that the expressiveness of the tree-based representations is greater, thus it is possible to map individuals encoded as \texttt{Linear} to trees, but not vice versa.
This property is used in one of our experiments.

\subsection{Opponent estimation}

Every evolution scheme evaluates individuals either by comparing how well they deal with a specified task, without a normalized score, or by using an external, predefined goal.
Both approaches have natural interpretations for CCGs -- a win rate against each other for the former and a win rate against a fixed opponent for the latter.

As the course of evolution using a predefined opponent will be heavily impacted by the opponent itself, two nontrivial questions arise.
What is the difference between using only the in-population evaluation from the one using a predefined opponent?
And what is the difference between using a weak and a strong opponent?
We have conducted two experiments to answer both of these questions in terms of the evolution process and resulting player's strength.

As both experiments required different evolution schemes, in total, three groups of individuals were evolved.
First one, called \texttt{progressive}, using the in-population evaluation (see Sec.~\ref{sec:repr_progressive}).
Second, called \texttt{weak-op}, using the existing \texttt{Baseline2} agent (\texttt{WeakOp}), one of the baseline LoCM agents.
During the draft phase, the agent chooses the creature card with the highest attack (if this is not possible it chooses the leftmost one).
During battle, it attacks everything on board starting with the cards with highest attack, aiming for the opponent first, and then opponent guard creatures starting from the highest health ones.
And third, called \texttt{strong-op}, using one of the pre-evolved \texttt{Tree-from-Linear} agents (\texttt{StrongOp}).
The last two are described in Sec.~\ref{sec:repr_op}.

\section{Representation Study:\\Trees Versus Vectors}

To measure the impact of different representations on the evolution, we ran the same experiment multiple times, each time substituting the underlying genome structure to one of described in Sec.~\ref{sec:repr_details}.
The metrics we find crucial are how well the agent performs in a real-life scenario, that is, in a proper tournament with other agents, and whether it progresses, i.e., plays better against own previous generations.

\begin{figure}[h]
  \centering
  \begin{small}
    \setlength{\size}{0.29\columnwidth}
    \setlength{\tabcolsep}{2pt}
    \begin{tabular}{cccc}
      {\rotatebox{90}{\parbox{\size}{\centering\texttt{weak-op}}}}&
      \includegraphics[angle=90,width=\size]{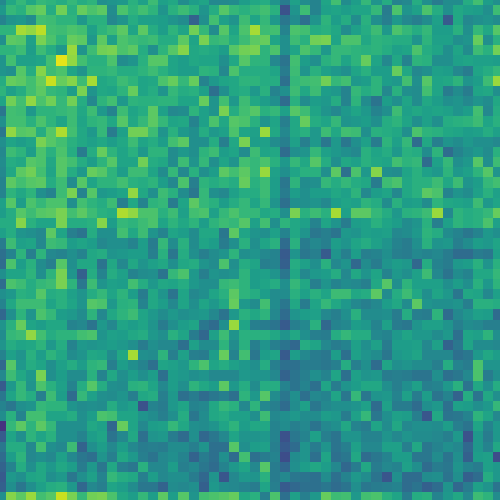}&
      \includegraphics[angle=90,width=\size]{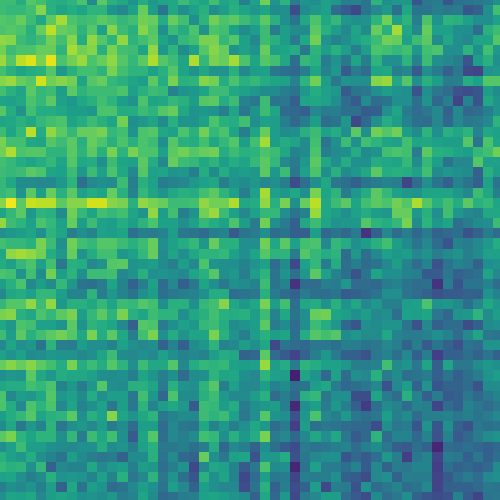}&
      \includegraphics[angle=90,width=\size]{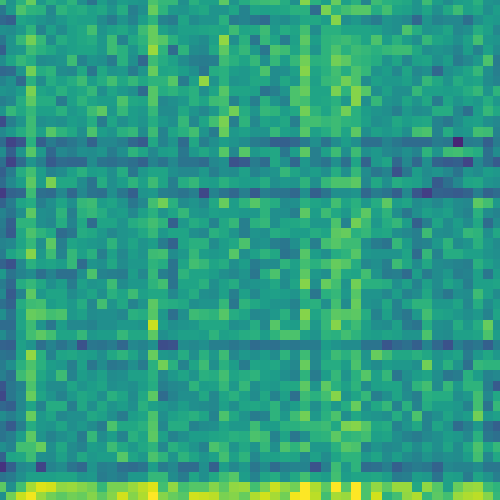}\\
      {\rotatebox{90}{\parbox{\size}{\centering\texttt{strong-op}}}}&
      \includegraphics[angle=90,width=\size]{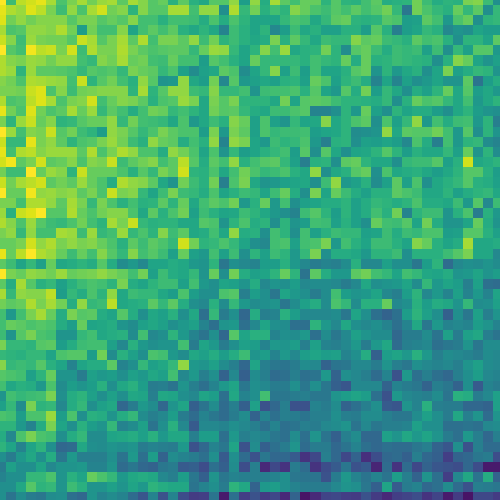}&
      \includegraphics[angle=90,width=\size]{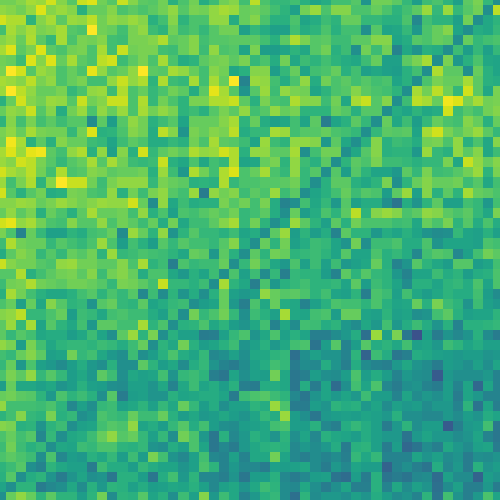}&
      \includegraphics[angle=90,width=\size]{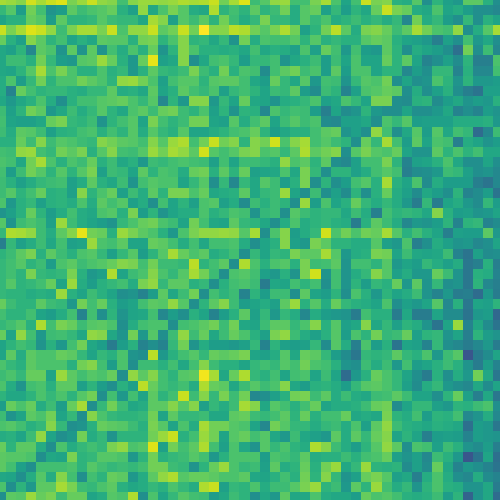}\\
      {\rotatebox{90}{\parbox{\size}{\centering\texttt{progressive}}}}&
      \includegraphics[angle=90,width=\size]{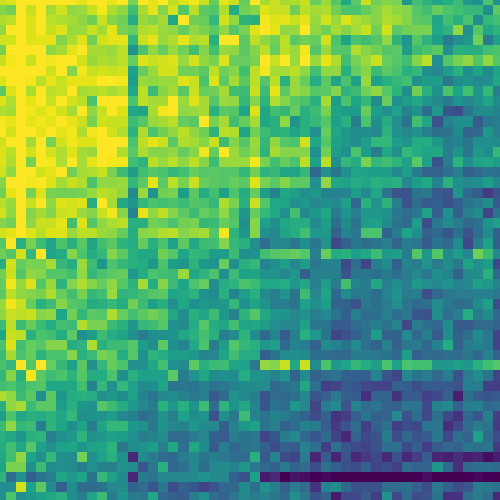}&
      \includegraphics[angle=90,width=\size]{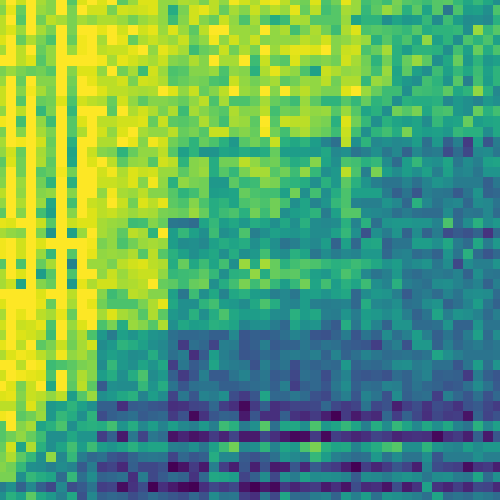}&
      \includegraphics[angle=90,width=\size]{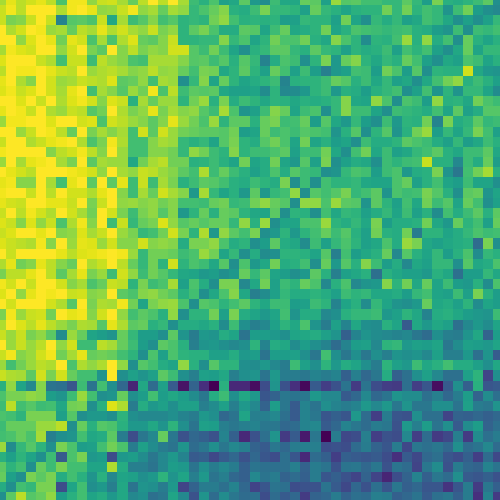}\\
      &
      \texttt{Linear}&
      \texttt{BinaryTree}&
      \texttt{Tree}
    \end{tabular}
  \end{small}
  \caption{
    Example self-play win rate heatmaps (other runs show similar properties).
    Each cell represents how well the best individual of generation on the y-axis plays against the best individual of generation on the x-axis.
    Visible diagonals present self-score (50\%).
    A highlighted triangle on the bottom left, present in all three \texttt{progressive} agents, proves that as the evolution progresses, so all individuals are increasingly better at self-play.
    Other heatmaps seem to be more random, indicating no clear progression in self-play.
  }
  \label{fig:heatmaps}
\end{figure}

\subsection{Experiment setup}\label{sec:repr_progressive}

We have evolved twelve copies of \texttt{progressive} agents, four for each representation.
Every run used the same parameters, that is 50 generations with a population of size 50 ($\mathit{population}$ parameter), elitism of 5 individuals, and the mutation rate of 5\%.
During the evaluation each two individuals played $\mathit{rounds}$ times on each of $\mathit{drafts}$ drafts on each side, which makes $2 \times (\mathit{population}-1) \times \mathit{drafts} \times \mathit{rounds}$ games in total.
In our experiments, $\mathit{drafts}=10$, and $\mathit{rounds}=10$.

In order to compare the agents in a standardized real-life scenario, we ran a tournament.
In addition to our evolved agents, we used two LoCM baseline agents -- \texttt{Baseline1} and \texttt{Baseline2} -- and four contestants of 2020 IEEE COG LoCM contest\footnote{\url{https://legendsofcodeandmagic.com/COG20/}} -- \texttt{Chad}, \texttt{Coac}, \texttt{OneLaneIsEnough}, and \texttt{ReinforcedGreediness}.

\subsection{Learning comparison}

As visible at the bottom row of Fig.~\ref{fig:heatmaps}, all representations successfully converged into a green triangle at the bottom left.

Such shape means that the following generations were not only preserving the already gained knowledge but also improving on each step.
Therefore, every representation can be evolved, playing against own previous generations.
Moreover, the progress of the \texttt{Linear} representation seems to be more stable, almost constant, while tree-based representations tend to improve by making larger but sporadic leaps.

This is not the case for the top and middle row, representing evolution with a predefined opponent (described in detail in the next section).
While the \texttt{Linear} representation manages to preserve hardly visible progress, both tree representations are more random, indicating no clear improvements in self-play.
Furthermore, the top row contains a few red stripes, indicating an exceptionally weak agent.
It is possible, as the evolution goal does not take self-play into consideration at all.

To measure how significant the learning progress is, we compare the win rate of the best individual of the first and the last generation against the best individuals of all generations.

\begin{figure}
  \centering
  \begin{tikzpicture}
    \begin{groupplot}[
      small,
      grid style=dashed,
      group style={
        group name=my plots,
        group size=1 by 3,
        horizontal sep=0pt,
        vertical sep=0pt,
        xlabels at=edge bottom,
        xticklabels at=edge bottom,
        ylabels at=edge left,
        yticklabels at=edge left,
      },
      height=165pt,
      legend cell align={right},
      legend style={
        anchor=south east,
        at={(1,0)},
        font=\small,
        row sep=-2pt,
      },
      width=\columnwidth,
      xlabel={Generation},
      xmax=50,
      xmin=0,
      xtick={0,10,20,30,40,50},
      ymajorgrids=true,
      ymax=75,
      ymin=35,
      ytick={35,40,45,50,55,60,65,70,75},
      yticklabels={,40,,50,,60,,70,},
      every axis plot/.append style={line cap=round, line join=round, very thick},
    ]
      \nextgroupplot[ylabel={\texttt{weak-op}}]
      \coordinate (top) at (rel axis cs:0,1);
      \addplot[color=red]coordinates {(0,32.85)(1,44.48)(2,47.33)(3,47.26)(4,41.76)(5,45.10)(6,50.72)(7,52.89)(8,53.55)(9,52.94)(10,53.13)(11,51.90)(12,51.27)(13,52.09)(14,50.80)(15,52.95)(16,51.22)(17,51.72)(18,51.35)(19,52.17)(20,55.09)(21,56.76)(22,55.06)(23,54.84)(24,52.03)(25,54.98)(26,54.64)(27,55.40)(28,56.97)(29,55.56)(30,55.86)(31,57.65)(32,60.57)(33,60.49)(34,58.14)(35,55.20)(36,59.64)(37,60.29)(38,61.25)(39,61.03)(40,60.76)(41,62.97)(42,59.13)(43,60.28)(44,58.98)(45,63.47)(46,59.43)(47,59.40)(48,60.29)(49,59.96)};
      \addplot[color=green]coordinates {(0,55.71)(1,42.20)(2,46.94)(3,46.93)(4,51.84)(5,50.69)(6,52.75)(7,57.96)(8,58.72)(9,58.40)(10,48.99)(11,46.35)(12,52.60)(13,53.31)(14,54.66)(15,53.37)(16,50.24)(17,46.37)(18,47.49)(19,51.23)(20,57.27)(21,56.83)(22,59.97)(23,50.49)(24,53.75)(25,55.06)(26,55.81)(27,55.21)(28,55.47)(29,62.24)(30,52.97)(31,57.21)(32,59.52)(33,60.12)(34,56.18)(35,59.23)(36,59.40)(37,57.10)(38,57.62)(39,57.11)(40,56.49)(41,59.88)(42,57.34)(43,59.84)(44,54.60)(45,59.52)(46,58.93)(47,60.09)(48,57.84)(49,57.00)};
      \addplot[color=blue]coordinates {(0,57.33)(1,65.81)(2,49.82)(3,56.07)(4,50.58)(5,61.57)(6,48.57)(7,50.10)(8,55.62)(9,49.12)(10,47.97)(11,51.32)(12,51.05)(13,52.17)(14,53.46)(15,57.74)(16,49.86)(17,53.64)(18,50.81)(19,53.16)(20,53.36)(21,54.93)(22,53.63)(23,54.75)(24,55.03)(25,53.87)(26,53.95)(27,56.95)(28,62.33)(29,61.81)(30,53.70)(31,61.30)(32,57.34)(33,58.99)(34,58.27)(35,56.79)(36,59.48)(37,60.57)(38,61.42)(39,57.90)(40,58.16)(41,59.50)(42,57.49)(43,59.24)(44,57.92)(45,58.99)(46,57.60)(47,56.68)(48,58.89)(49,57.46)};
      \nextgroupplot[ylabel={\texttt{strong-op}}]
      \addplot[color=red]coordinates {(0,40.71)(1,48.31)(2,47.16)(3,46.01)(4,48.90)(5,48.21)(6,51.79)(7,51.23)(8,52.06)(9,51.35)(10,53.25)(11,52.46)(12,52.96)(13,52.67)(14,53.89)(15,54.69)(16,54.09)(17,53.83)(18,54.62)(19,56.27)(20,56.81)(21,55.52)(22,57.05)(23,57.22)(24,57.49)(25,59.41)(26,59.67)(27,58.15)(28,58.51)(29,60.59)(30,61.33)(31,60.77)(32,60.56)(33,60.46)(34,59.69)(35,60.69)(36,62.13)(37,57.64)(38,60.68)(39,59.38)(40,63.41)(41,63.86)(42,64.65)(43,63.12)(44,62.68)(45,64.12)(46,63.59)(47,64.78)(48,64.68)(49,66.15)};
      \addplot[color=green]coordinates {(0,56.11)(1,52.82)(2,52.23)(3,52.02)(4,52.10)(5,54.29)(6,52.40)(7,48.37)(8,52.53)(9,54.69)(10,55.97)(11,55.14)(12,52.44)(13,49.80)(14,54.77)(15,55.40)(16,54.78)(17,56.13)(18,55.42)(19,55.60)(20,57.12)(21,55.42)(22,57.65)(23,57.77)(24,57.94)(25,57.45)(26,56.99)(27,59.11)(28,60.09)(29,60.74)(30,60.39)(31,61.06)(32,60.43)(33,62.23)(34,62.09)(35,56.97)(36,65.08)(37,64.42)(38,64.73)(39,65.40)(40,65.16)(41,65.62)(42,65.26)(43,64.74)(44,64.05)(45,65.39)(46,64.81)(47,64.65)(48,65.77)(49,65.18)};
      \addplot[color=blue]coordinates {(0,57.07)(1,56.72)(2,56.27)(3,56.75)(4,57.00)(5,58.04)(6,55.59)(7,56.52)(8,57.00)(9,56.76)(10,58.27)(11,57.83)(12,58.39)(13,57.95)(14,57.36)(15,56.45)(16,57.06)(17,57.40)(18,56.45)(19,57.59)(20,55.42)(21,54.97)(22,56.36)(23,56.19)(24,56.93)(25,56.78)(26,60.67)(27,58.34)(28,59.78)(29,61.41)(30,61.82)(31,60.19)(32,60.25)(33,61.42)(34,62.21)(35,62.36)(36,61.68)(37,62.44)(38,61.09)(39,61.09)(40,62.87)(41,64.27)(42,62.94)(43,62.04)(44,62.53)(45,63.55)(46,67.64)(47,65.18)(48,64.05)(49,65.55)};
      \nextgroupplot[ylabel={\texttt{progressive}}]
      \addplot[color=red]coordinates {(0,35.36)(1,36.62)(2,36.71)(3,42.31)(4,43.49)(5,45.30)(6,46.02)(7,48.01)(8,47.08)(9,50.29)(10,48.95)(11,49.81)(12,52.36)(13,54.95)(14,54.40)(15,53.97)(16,52.37)(17,57.57)(18,56.83)(19,58.61)(20,58.76)(21,58.70)(22,55.04)(23,57.28)(24,60.17)(25,58.94)(26,58.27)(27,59.98)(28,59.29)(29,60.26)(30,60.88)(31,61.10)(32,61.64)(33,63.48)(34,63.55)(35,62.31)(36,64.48)(37,61.27)(38,64.88)(39,64.87)(40,63.71)(41,65.66)(42,63.37)(43,65.16)(44,65.88)(45,66.37)(46,65.60)(47,65.79)(48,66.21)(49,66.62)};\addlegendentry{\texttt{Linear}};
      \addplot[color=green]coordinates {(0,46.74)(1,39.50)(2,44.21)(3,45.55)(4,48.81)(5,50.89)(6,43.27)(7,48.74)(8,46.84)(9,47.30)(10,47.70)(11,50.81)(12,48.98)(13,49.24)(14,48.83)(15,51.90)(16,45.99)(17,57.84)(18,58.35)(19,58.04)(20,57.32)(21,60.50)(22,61.61)(23,59.35)(24,62.17)(25,61.76)(26,58.91)(27,61.01)(28,63.04)(29,61.77)(30,63.36)(31,62.43)(32,65.12)(33,64.47)(34,63.44)(35,65.76)(36,67.43)(37,66.77)(38,64.57)(39,66.99)(40,69.30)(41,68.34)(42,68.02)(43,68.95)(44,68.90)(45,68.55)(46,69.61)(47,66.01)(48,67.93)(49,66.86)};\addlegendentry{\texttt{BinaryTree}};
      \addplot[color=blue]coordinates {(0,46.71)(1,49.05)(2,47.27)(3,51.79)(4,50.04)(5,53.02)(6,52.60)(7,55.03)(8,50.79)(9,54.54)(10,53.74)(11,51.79)(12,57.12)(13,59.30)(14,58.28)(15,59.05)(16,57.18)(17,60.36)(18,59.40)(19,60.47)(20,61.30)(21,60.96)(22,61.03)(23,61.25)(24,62.19)(25,62.19)(26,61.70)(27,62.79)(28,62.61)(29,63.44)(30,61.71)(31,61.16)(32,61.15)(33,63.49)(34,61.49)(35,62.41)(36,64.81)(37,63.51)(38,62.83)(39,65.70)(40,63.51)(41,65.16)(42,65.96)(43,63.45)(44,63.38)(45,64.17)(46,63.31)(47,63.66)(48,65.96)(49,66.32)};\addlegendentry{\texttt{Tree}};
      \coordinate (bot) at (rel axis cs:1,0);
    \end{groupplot}
    \path (top-|current bounding box.west)--node[anchor=south,rotate=90] {\small{Win rate (\%)}}(bot-|current bounding box.west);
  \end{tikzpicture}
  \caption{
    Evolution progress of all the agents.
    Best individuals from a generation (x-axis) fought against the top individuals of all own generations, yielding an average win rate (y-axis).
  }
  \label{fig:evolution}
\end{figure}
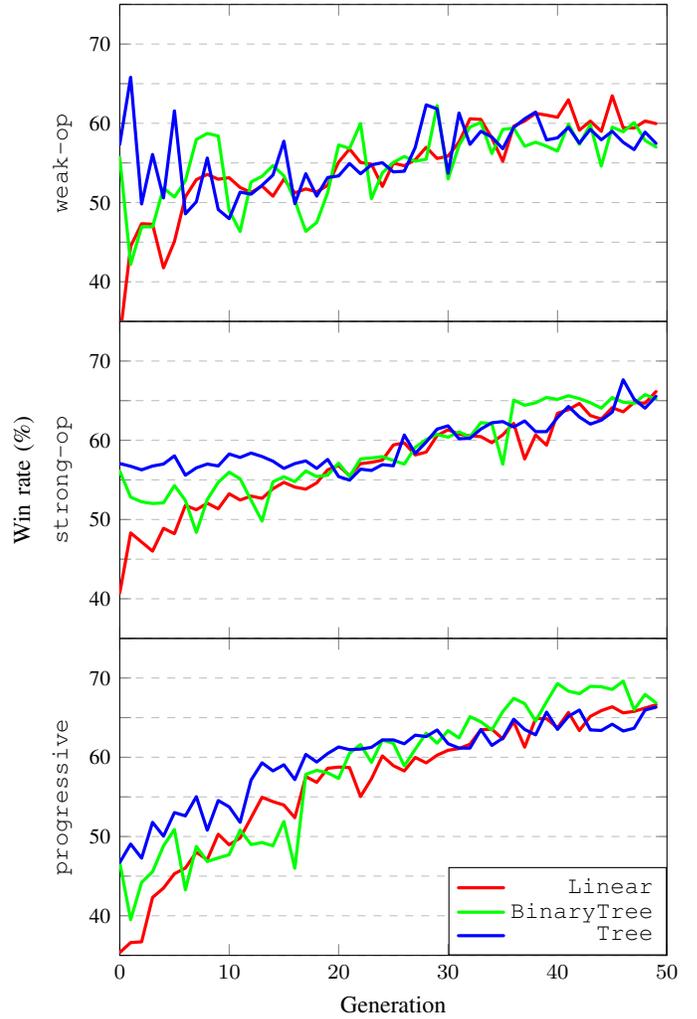

With such a metric in mind, the \texttt{Linear} representation stands out again.
As presented in Fig.~\ref{fig:evolution}, both \texttt{BinaryTree} and \texttt{Tree} result in a smaller difference of about 20\% whereas the \texttt{Linear} representation achieves over 31\% difference on average, across all the evolution schemes.

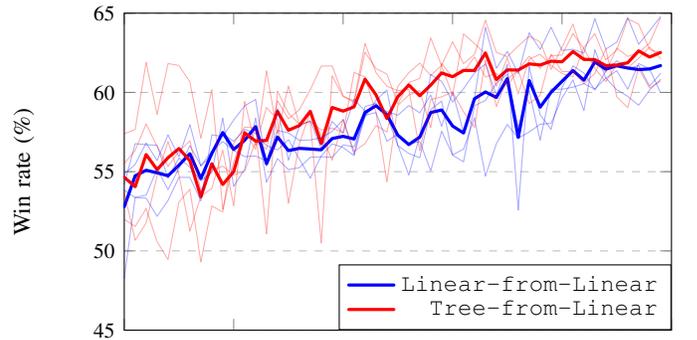
\begin{figure}
  \centering
  \begin{tikzpicture}
    \begin{groupplot}[
      small,
      grid style=dashed,
      group style={
        group name=my plots,
        group size=1 by 3,
        horizontal sep=0pt,
        vertical sep=0pt,
        xlabels at=edge bottom,
        xticklabels at=edge bottom,
        ylabels at=edge left,
        yticklabels at=edge left,
      },
      height=165pt,
      legend cell align={right},
      legend style={
        anchor=south east,
        at={(1,0)},
        font=\small,
        row sep=-2pt,
      },
      width=\columnwidth,
      xlabel={Generation},
      xmax=50,
      xmin=0,
      xtick={0,10,20,30,40,50},
      ymajorgrids=true,
      ymax=65,
      ymin=45,
      ytick={45,50,55,60,65},
      yticklabels={45,50,55,60,65},
      every axis plot/.append style={line cap=round, line join=round},
    ]
      \nextgroupplot[ylabel={\color{white}\texttt{|}}]
      \coordinate (top) at (rel axis cs:0,1);
      \addplot[color=blue, style=very thick]coordinates {(0,52.78)(1,54.74)(2,55.09)(3,54.93)(4,54.74)(5,55.41)(6,56.13)(7,54.57)(8,56.15)(9,57.47)(10,56.40)(11,56.98)(12,57.84)(13,55.50)(14,57.19)(15,56.33)(16,56.47)(17,56.43)(18,56.38)(19,57.09)(20,57.23)(21,57.06)(22,58.73)(23,59.16)(24,58.60)(25,57.30)(26,56.71)(27,57.18)(28,58.73)(29,58.89)(30,57.88)(31,57.44)(32,59.60)(33,60.03)(34,59.68)(35,60.87)(36,57.16)(37,60.75)(38,59.07)(39,60.01)(40,60.70)(41,61.38)(42,60.77)(43,61.92)(44,61.47)(45,61.68)(46,61.54)(47,61.44)(48,61.48)(49,61.69)};\addlegendentry{\texttt{Linear-from-Linear}};
      \addplot[color=red, style=very thick]coordinates {(0,54.65)(1,54.05)(2,56.08)(3,55.13)(4,55.88)(5,56.46)(6,55.69)(7,53.41)(8,55.50)(9,54.19)(10,55.00)(11,57.45)(12,56.93)(13,56.97)(14,58.81)(15,57.61)(16,57.91)(17,58.81)(18,56.77)(19,59.05)(20,58.82)(21,59.09)(22,60.84)(23,59.85)(24,58.35)(25,59.73)(26,60.47)(27,59.79)(28,60.46)(29,61.23)(30,60.99)(31,61.39)(32,61.38)(33,62.49)(34,60.81)(35,61.42)(36,61.41)(37,61.80)(38,61.73)(39,61.97)(40,61.93)(41,62.58)(42,62.08)(43,62.07)(44,61.69)(45,61.73)(46,61.87)(47,62.62)(48,62.23)(49,62.52)};\addlegendentry{\texttt{Tree-from-Linear}};

      \addplot[color=blue, draw opacity=0.33]coordinates {(0,53.84)(1,53.36)(2,53.33)(3,52.18)(4,53.18)(5,54.64)(6,54.64)(7,53.16)(8,57.33)(9,57.21)(10,58.26)(11,59.15)(12,59.56)(13,55.37)(14,58.98)(15,55.73)(16,57.86)(17,58.22)(18,56.44)(19,58.17)(20,57.43)(21,59.58)(22,58.82)(23,59.11)(24,58.80)(25,58.56)(26,59.17)(27,58.71)(28,59.84)(29,57.86)(30,60.03)(31,57.49)(32,59.62)(33,64.10)(34,61.53)(35,62.59)(36,61.18)(37,63.76)(38,62.71)(39,64.11)(40,62.65)(41,62.72)(42,62.48)(43,64.65)(44,62.73)(45,63.17)(46,64.73)(47,63.04)(48,63.89)(49,64.78)};
      \addplot[color=blue, draw opacity=0.33]coordinates {(0,48.26)(1,53.38)(2,53.51)(3,55.17)(4,53.87)(5,55.21)(6,56.64)(7,55.37)(8,56.29)(9,56.90)(10,56.34)(11,55.95)(12,55.99)(13,55.67)(14,57.60)(15,57.44)(16,56.98)(17,57.00)(18,57.14)(19,57.54)(20,58.28)(21,56.10)(22,59.11)(23,59.31)(24,58.54)(25,56.65)(26,54.79)(27,57.25)(28,61.13)(29,60.70)(30,59.39)(31,60.23)(32,61.23)(33,61.80)(34,59.83)(35,61.77)(36,52.58)(37,61.65)(38,58.59)(39,57.67)(40,60.30)(41,61.68)(42,61.22)(43,60.15)(44,61.34)(45,62.00)(46,61.48)(47,61.70)(48,60.40)(49,61.19)};
      \addplot[color=blue, draw opacity=0.33]coordinates {(0,54.84)(1,55.46)(2,56.81)(3,55.70)(4,57.32)(5,55.94)(6,57.97)(7,55.36)(8,56.03)(9,57.39)(10,56.92)(11,57.67)(12,57.43)(13,56.64)(14,56.37)(15,57.02)(16,55.80)(17,54.59)(18,55.99)(19,56.75)(20,56.53)(21,56.01)(22,58.56)(23,58.73)(24,56.92)(25,57.37)(26,56.75)(27,56.90)(28,55.05)(29,57.55)(30,56.06)(31,56.42)(32,58.40)(33,54.72)(34,56.76)(35,57.98)(36,57.11)(37,57.50)(38,57.99)(39,58.05)(40,59.65)(41,60.79)(42,58.85)(43,62.69)(44,60.90)(45,61.37)(46,60.20)(47,61.82)(48,61.48)(49,59.98)};
      \addplot[color=blue, draw opacity=0.33]coordinates {(0,54.19)(1,56.77)(2,56.70)(3,56.67)(4,54.59)(5,55.86)(6,55.29)(7,54.42)(8,54.94)(9,58.38)(10,54.07)(11,55.15)(12,58.38)(13,54.34)(14,55.82)(15,55.13)(16,55.25)(17,55.91)(18,55.94)(19,55.90)(20,56.69)(21,56.56)(22,58.45)(23,59.50)(24,60.15)(25,56.63)(26,56.15)(27,55.88)(28,58.92)(29,59.46)(30,56.04)(31,55.62)(32,59.16)(33,59.49)(34,60.60)(35,61.15)(36,57.80)(37,60.11)(38,56.99)(39,60.22)(40,60.20)(41,60.33)(42,60.55)(43,60.21)(44,60.91)(45,60.17)(46,59.75)(47,59.19)(48,60.15)(49,60.82)};

      \addplot[color=red, draw opacity=0.33]coordinates {(0,57.39)(1,57.66)(2,61.90)(3,58.62)(4,61.82)(5,61.70)(6,60.66)(7,57.11)(8,60.17)(9,52.62)(10,55.59)(11,57.57)(12,56.42)(13,61.36)(14,57.38)(15,60.40)(16,57.97)(17,58.16)(18,57.63)(19,57.75)(20,57.46)(21,61.26)(22,60.97)(23,58.74)(24,54.36)(25,59.18)(26,59.83)(27,60.11)(28,57.78)(29,61.83)(30,60.95)(31,61.29)(32,62.23)(33,60.98)(34,61.20)(35,60.39)(36,61.42)(37,62.28)(38,61.51)(39,62.31)(40,62.36)(41,63.71)(42,62.76)(43,61.51)(44,60.68)(45,61.75)(46,61.54)(47,62.15)(48,62.86)(49,62.15)};
      \addplot[color=red, draw opacity=0.33]coordinates {(0,53.69)(1,50.67)(2,51.76)(3,54.73)(4,56.90)(5,54.80)(6,51.25)(7,53.45)(8,53.94)(9,56.35)(10,52.85)(11,59.43)(12,56.65)(13,59.04)(14,58.30)(15,53.01)(16,56.78)(17,58.76)(18,57.29)(19,55.73)(20,58.31)(21,56.27)(22,56.77)(23,57.91)(24,57.50)(25,60.39)(26,61.24)(27,61.85)(28,61.07)(29,61.71)(30,63.20)(31,61.82)(32,62.40)(33,64.55)(34,62.53)(35,62.25)(36,62.93)(37,63.54)(38,63.19)(39,61.45)(40,62.85)(41,64.31)(42,62.20)(43,61.87)(44,61.28)(45,62.80)(46,62.27)(47,62.51)(48,63.11)(49,64.70)};
      \addplot[color=red, draw opacity=0.33]coordinates {(0,51.99)(1,51.54)(2,52.67)(3,50.58)(4,49.46)(5,53.09)(6,53.89)(7,49.30)(8,52.84)(9,52.45)(10,56.78)(11,57.81)(12,57.76)(13,51.07)(14,60.08)(15,58.69)(16,58.00)(17,58.16)(18,50.50)(19,61.04)(20,59.73)(21,59.47)(22,63.08)(23,61.25)(24,60.39)(25,60.11)(26,60.20)(27,56.32)(28,59.55)(29,59.72)(30,60.02)(31,58.78)(32,57.80)(33,62.02)(34,59.11)(35,60.85)(36,61.39)(37,59.30)(38,59.82)(39,62.41)(40,61.31)(41,61.32)(42,60.55)(43,61.20)(44,62.74)(45,59.72)(46,59.04)(47,61.52)(48,60.27)(49,60.72)};
      \addplot[color=red, draw opacity=0.33]coordinates {(0,55.51)(1,56.34)(2,58.01)(3,56.59)(4,55.33)(5,56.27)(6,56.95)(7,53.78)(8,55.07)(9,55.35)(10,54.79)(11,55.01)(12,56.89)(13,56.42)(14,59.48)(15,58.33)(16,58.91)(17,60.17)(18,61.66)(19,61.67)(20,59.79)(21,59.37)(22,62.53)(23,61.51)(24,61.16)(25,59.26)(26,60.62)(27,60.88)(28,63.44)(29,61.68)(30,59.79)(31,63.69)(32,63.09)(33,62.42)(34,60.39)(35,62.19)(36,59.91)(37,62.09)(38,62.41)(39,61.71)(40,61.19)(41,60.99)(42,62.84)(43,63.70)(44,62.05)(45,62.68)(46,64.64)(47,64.31)(48,62.70)(49,62.50)};
      \coordinate (bot) at (rel axis cs:1,0);
    \end{groupplot}
    \path (top-|current bounding box.west)--node[anchor=south,rotate=90] {\small{Win rate (\%)}}(bot-|current bounding box.west);
  \end{tikzpicture}
  \caption{
    Evolution progress of the \texttt{from-Linear} agents.
    Best individuals from a generation (x-axis) fought against the top individuals of all own generations, yielding an average win rate (y-axis).
    Each of the four \texttt{Linear} agents was used as a base for the initial population twice.
    The two bold lines average the thin, semi-transparent lines that are the averaged results of agents with the same base.
  }
  \label{fig:from-linear}
\end{figure}


\subsection{Tournament comparison}

When evolved using a fixed, weak opponent (\texttt{weak-op}), the difference between the representations is significant.
There is a huge, almost 10\% wide, gap between \texttt{Linear} (42.9\% average win rate) and \texttt{BinaryTree} (33.8\%).
The \texttt{Tree} representation is in between, performing slightly better than its binary counterpart and achieving 36.2\%.

\newpage

Results are similar when evolved using self-play evaluation and a randomly initialized population (\texttt{progressive}).
The \texttt{Linear} representation again yields strictly better results in the tournament (54.8\%) than both \texttt{BinaryTree} and \texttt{Tree} (45.6\% and 45.7\% respectively).

However, using a better opponent (\texttt{strong-op}) yields completely different results.
In this scenario, the differences between agents' performance are less significant, around 4\%.
To be precise, \texttt{Linear} achieved almost 49\%, \texttt{BinaryTree} slightly over 45\%, and \texttt{Tree} nearly 45\% average win rate.

This matches the results of playing against own previous generations, described in the previous subsection.
We conclude that it is an implication of the capacity of the representation.

While a more limited \texttt{Linear} representation finds a decent solution sooner and gradually improves it, more capable tree-based representations regularly leap towards the goal.

More detailed tournament results, limited to our agents and their evolutionary targets, are presented in Tables \ref{table:tournament-weak-op}, \ref{table:tournament-strong-op}, and \ref{table:tournament-progressive}.

\renewcommand{\arraystretch}{1.05}
\begin{table*}[!b]
  \centering
  \caption{
    \centering
    Subset of the tournament results, presenting all \texttt{weak-op} agents and their evolution target, \texttt{WeakOp}, itself.
    \newline
    Every score is an average win rate (along with its standard deviation) of the agent on the left against the agent on top.
    \newline
    Given average win rate is calculated based on the full tournament results.
  }
  \begin{small}
    \begin{tabular}{r|cccc|c}
                          & \texttt{Linear} & \texttt{BinaryTree} & \texttt{Tree} & \texttt{WeakOp} & Global avg.   \\
      \hline
          \texttt{Linear} & --                & 56.9$\pm$10.2\%       & 47.5$\pm$12.6\% & 65.9$\pm$10.0\%   & 42.9$\pm$15.7\% \\
      \texttt{BinaryTree} & 43.0$\pm$10.2\%   & --                    & 49.2$\pm$12.0\% & 82.9$\pm$2.67\%   & 33.8$\pm$14.2\% \\
            \texttt{Tree} & 52.4$\pm$12.6\%   & 50.7$\pm$12.0\%       & --              & 56.0$\pm$8.25\%   & 36.2$\pm$15.7\% \\
          \texttt{WeakOp} & 34.0$\pm$10.0\%   & 17.0$\pm$2.67\%       & 43.9$\pm$8.25\% & --                & 42.8$\pm$17.3\%
    \end{tabular}
  \end{small}
  \label{table:tournament-weak-op}
\end{table*}

\begin{table*}[!b]
  \centering
  \caption{
    \centering
    Subset of the tournament results, presenting all \texttt{strong-op} agents and their evolution target, \texttt{StrongOp}, itself.
    \newline
    Every score is an average win rate (along with its standard deviation) of the agent on the left against the agent on top.
    \newline
    Given average win rate is calculated based on the full tournament results.
  }
  \begin{small}
    \begin{tabular}{r|cccc|c}
                          & \texttt{Linear} & \texttt{BinaryTree} & \texttt{Tree} & \texttt{StrongOp} & Global avg.   \\
      \hline
          \texttt{Linear} & --                & 52.1$\pm$13.3\%       & 52.9$\pm$9.15\% & 38.8$\pm$11.5\%     & 48.9$\pm$16.6\% \\
      \texttt{BinaryTree} & 47.8$\pm$13.3\%   & --                    & 49.9$\pm$11.7\% & 34.1$\pm$9.90\%     & 45.1$\pm$16.9\% \\
            \texttt{Tree} & 47.1$\pm$9.15\%   & 50.0$\pm$11.7\%       & --              & 36.7$\pm$9.51\%     & 44.8$\pm$15.7\% \\
        \texttt{StrongOp} & 61.1$\pm$11.5\%   & 65.8$\pm$9.90\%       & 63.2$\pm$9.51\% & --                  & 62.8$\pm$14.6\%
    \end{tabular}
  \end{small}
  \label{table:tournament-strong-op}
\end{table*}

\begin{table*}[!b]
  \centering
  \caption{
    \centering
    Subset of the tournament results, presenting all \texttt{progressive} and \texttt{from-Linear} agents.
    \newline
    Every score is an average win rate (along with its standard deviation) of the agent on the left against the agent on top.
    \newline
    Given average win rate is calculated based on the full tournament results.
  }
  \begin{small}
    \begin{tabular}{r|ccccc|c}
                                  & \texttt{Linear} & \texttt{BinaryTree} & \texttt{Tree} & \texttt{Linear-from-Linear} & \texttt{Tree-from-Linear} & Global avg.   \\
      \hline
                  \texttt{Linear} & --                & 59.7$\pm$14.2\%       & 61.2$\pm$14.1\% & 43.5$\pm$8.40\%               & 41.0$\pm$13.7\%             & 54.8$\pm$18.0\% \\
              \texttt{BinaryTree} & 40.2$\pm$14.2\%   & --                    & 50.1$\pm$13.1\% & 40.0$\pm$11.5\%               & 37.6$\pm$10.3\%             & 45.5$\pm$16.8\% \\
                    \texttt{Tree} & 38.7$\pm$14.1\%   & 49.8$\pm$13.1\%       & --              & 39.5$\pm$15.7\%               & 35.2$\pm$13.5\%             & 45.6$\pm$18.7\% \\
      \texttt{Linear-from-Linear} & 56.4$\pm$8.40\%   & 60.0$\pm$11.5\%       & 60.4$\pm$15.7\% & --                            & 44.3$\pm$11.1\%             & 58.0$\pm$17.1\% \\
        \texttt{Tree-from-Linear} & 58.9$\pm$13.7\%   & 62.3$\pm$10.3\%       & 64.7$\pm$13.5\% & 55.6$\pm$11.1\%               & --                          & 60.2$\pm$17.4\%
    \end{tabular}
  \end{small}
  \label{table:tournament-progressive}
\end{table*}

\subsection{Tuning good solutions}

Knowing that the tree-based model is more general but also harder to learn, the natural question is if we can use it to improve the solutions, instead of generating them from scratch.
The ideal scenario will be to reach a limit of optimization based on the linear representation, encode obtained solutions into the tree format, and continue evolution using this stronger model.

To verify this hypothesis, we have evolved two more agents: \texttt{Linear-from-Linear} and \texttt{Tree-from-Linear}.
Each of the four previously evolved \texttt{Linear-progressive} agents was used as a base for a second evolution process.

Now, instead of randomly, the population was initialized with copies of the base agent, each one mutated $\mathit{n}=5$ times; hence the \texttt{from-Linear} suffix in their name.

However, translation of a \texttt{Linear} representation into a \texttt{Tree} representation is ambiguous.
We have used what we believe is the most straightforward one -- an \texttt{Add} operator in the root with a list of \texttt{Mul} + \texttt{Literal} + \texttt{Feature} subtrees, one for each of the available features.

As expected, the tournament results for such pre-evolved agents are entirely different.
For the first time, the \texttt{Linear} representation is not the top one.
The \texttt{Tree} agent performs better, ending up with an average win rate of over 60\%, whereas the \texttt{Linear} agent finished with 58\%.
This is a relatively minor but consistent improvement that may further improve with a longer evolution.

A similar difference is visible in the process of evolution.
Both representations perform similarly, but \texttt{Tree} is on average above the \texttt{Linear} almost constantly.
Comparison of the best individuals across the generations is presented in Fig.~\ref{fig:from-linear}.

Our conclusion is that both results are implications of the representation.
While the more restricted \texttt{Linear} is not able to progress after a certain point, more expressive \texttt{Tree} benefits from the bootstrap and keeps improving.
Once again, more detailed results are presented in Table~\ref{table:tournament-progressive}.

\section{Opponent Estimation Study:\\Progressive Versus Fixed}

On one hand, a decent agent -- better than a fully random one -- usually emerges from the task definition itself.
It is often heuristic, filled with expert knowledge about the problem.
On the other, a standard in-population evaluation is proved to yield a good solution, e.g., using a linear combination of some expert-based features and a basic evolution scheme.
Our question is whether one of these approaches is superior to the other in terms of real-world evaluation.

Additionally, we are interested in what the difference is between using a weak and a strong opponent as a measure.
This is often an interesting dilemma, as both approaches may be potentially vulnerable to fast stagnation.
A weak opponent may be too easy to beat, so the win rate quickly approaches large values, while against a strong opponent, we may struggle to achieve any victories thus, there may be no reasonable progress at all.
(Of course, one may try evaluations based on a portfolio of agents, but this leads to other potential problems and is computationally much more expensive.)

\subsection{Experiment setup}\label{sec:repr_op}

To properly compare the two evolution schemes, one needs to compare not only their outcomes but also the costs.
In our case, the overhead of a given scheme is negligible in comparison to the cost of simulations.
Thus, we use the number of simulated games as a metric of evolution cost.
For the sake of simplicity, we assume that every game takes the same amount of time.
That is not true, as better agents tend to play longer.
In practice, the average time of a single game across the whole evolution run is comparable between schemes.

In \texttt{progressive} scenario, each two individuals played $\mathit{rounds}$ games on each of $\mathit{drafts}$ drafts.
In \texttt{weak-op} and \texttt{strong-op} scenarios, every individual played $\mathit{population} \times \mathit{rounds}$ games on each of $\mathit{drafts}$ drafts for each side.
Overall, all three evaluation schemes use the same number of games for each individual.
In our experiments, $\mathit{drafts}=10$, $\mathit{population}=50$, and $\mathit{rounds}=10$.

\subsection{Learning comparison}

To verify whether the evolution progresses, we compare the best individuals of all generations after the evolution finishes.
Such progress, visualized in Fig.~\ref{fig:heatmaps}, is definite in all of the \texttt{progressive} individuals and less significant for the opponent-based ones.
It is clear that the \texttt{progressive} scheme is better at this task -- the elitism in combination with the in-population evaluation implies it.

Additionally, all three heatmaps of \texttt{weak-op} agents have some bold red stripes.
Every stripe represents a few consecutive agents that were significantly weaker than the local average.
The same happens in other evolution schemes, but it is rather exceptional.
It is understandable, as only the \texttt{progressive} evolution takes in-population evaluation into account.
But also, it shows that learning against the stronger opponent is more resilient to the forgetting issue.

\subsection{Tournament comparison}

The lack of progress visible in heatmaps does not imply a lack of general improvement.
As seen in Table~\ref{table:tournament-strong-op} and Table~\ref{table:tournament-progressive}, the two tree-based representations achieve comparable results for both \texttt{progressive} and \texttt{strong-op} -- around 45\%.

However, this does not hold for the most straightforward \texttt{Linear} representation.
The differences between \texttt{weak-op}, \texttt{strong-op}, and \texttt{progressive} are large, around 6\% each. 

When we compare how well both fixed-opponent agents play against their evolution goals, we see that their scores do not correlate with the representation.
The \texttt{BinaryTree} representation performs best in the \texttt{weak-op} variant and achieved 82.9\% wins against the \texttt{WeakOp}, while the other representations were much weaker -- 65.9\% and 56\% for \texttt{Linear} and \texttt{Tree} respectively.
At the same time, \texttt{BinaryTree} is the worst in the \texttt{strong-op} variant, achieving only 34.1\% wins against \texttt{StrongOp}, whereas \texttt{Linear} and \texttt{Tree} achieved 38.8\% and 36.7\% wins respectively.

\begin{figure*}
  \centering
  \begin{tikzpicture}
    \definecolor{color1}{RGB}{255,191,0}
    \definecolor{color2}{RGB}{0,255,64}
    \definecolor{color3}{RGB}{0,64,255}
    \definecolor{color4}{RGB}{255,0,191}
    \begin{axis}[
      small,
      grid style=dashed,
      height=187pt,
      legend cell align={right},
      legend style={
        anchor=west,
        at={(1.025,0.5)},
        font=\small,
        row sep=1.5pt,
      },
      line cap=round,
      line join=round,
      mark repeat=20,
      width=1.5\columnwidth,
      xlabel={Round},
      xmax=100,
      xmin=0,
      xtick={0,100},
      xticklabels={,,},
      ylabel={Win rate (\%)},
      ymajorgrids=true,
      ymax=65,
      ymin=30,
      ytick={30,35,40,45,50,55,60,65},
      every axis plot/.append style={very thick},
    ]
      \addplot[color1, mark=        *] table[col sep=semicolon, x expr=\coordindex, y index= 7] {graph.data};\addlegendentry{\texttt{Linear-weak-op}}
      \addplot[color1, mark=triangle*] table[col sep=semicolon, x expr=\coordindex, y index= 2] {graph.data};\addlegendentry{\texttt{BinaryTree-weak-op}}
      \addplot[color1, mark=  square*] table[col sep=semicolon, x expr=\coordindex, y index=13] {graph.data};\addlegendentry{\texttt{Tree-weak-op}}
      \addplot[color2, mark=        *] table[col sep=semicolon, x expr=\coordindex, y index= 9] {graph.data};\addlegendentry{\texttt{Linear-strong-op}}
      \addplot[color2, mark=triangle*] table[col sep=semicolon, x expr=\coordindex, y index= 3] {graph.data};\addlegendentry{\texttt{BinaryTree-strong-op}}
      \addplot[color2, mark=  square*] table[col sep=semicolon, x expr=\coordindex, y index=15] {graph.data};\addlegendentry{\texttt{Tree-strong-op}}
      \addplot[color3, mark=        *] table[col sep=semicolon, x expr=\coordindex, y index=10] {graph.data};\addlegendentry{\texttt{Linear-progressive}}
      \addplot[color3, mark=triangle*] table[col sep=semicolon, x expr=\coordindex, y index= 4] {graph.data};\addlegendentry{\texttt{BinaryTree-progressive}}
      \addplot[color3, mark=  square*] table[col sep=semicolon, x expr=\coordindex, y index=16] {graph.data};\addlegendentry{\texttt{Tree-progressive}}
      \addplot[color4, mark=        *] table[col sep=semicolon, x expr=\coordindex, y index= 8] {graph.data};\addlegendentry{\texttt{Linear-from-Linear}}
      \addplot[color4, mark=  square*] table[col sep=semicolon, x expr=\coordindex, y index=14] {graph.data};\addlegendentry{\texttt{Tree-from-Linear}}
    \end{axis}
  \end{tikzpicture}
  \caption{
    A subset of the tournament results.
    All scores (y-axis) stabilize as the number of rounds (x-axis) increases.
  }
  \label{fig:tournament}
\end{figure*}
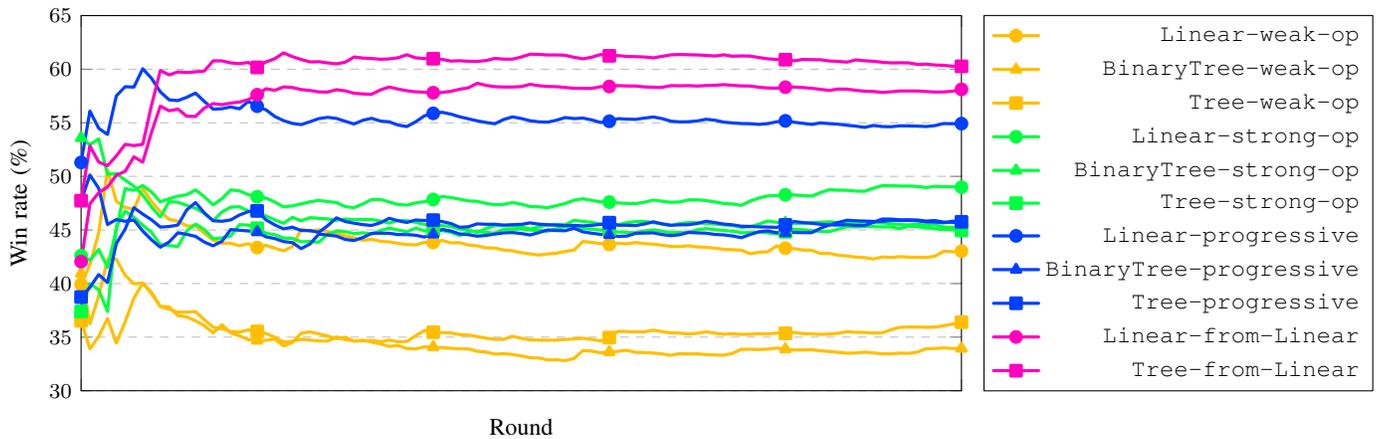

As presented in Fig.~\ref{fig:tournament}, all \texttt{weak-op} agents are strictly worse than the \texttt{strong-op} since almost the beginning of the tournament.
It is not the case for all of the \texttt{progressive} and \texttt{strong-op} agents -- every representation is indeed stronger in the former variant, but the \texttt{Linear-strong-op} agent is superior to both tree-like representations using \texttt{progressive} evolution.

To summarize, evolution via self-play yields better agents than evolution towards a fixed opponent.
However, it is not true if the representation is not fixed, e.g., \texttt{Tree-progressive} is inferior to \texttt{Linear-strong-op}.

Additionally, evolution using a fixed opponent has a slightly different cost characteristic.
While evolution using self-play simulates longer games gradually, using a fixed opponent starts with much shorter ones but ends with longer ones -- already trained agent quickly deals with initial, almost random agents and holds better against the evolved ones.
As expected, the \texttt{weak-op} evolution is faster than the \texttt{strong-op}.

\section{Conclusion}

In this work, we presented a study regarding two important aspects of evolving feature-based game evaluation functions: the choice of genome representation (implying the algorithm used) and the choice of opponent used to test the model.

Although our research was focused on the domain of collectible card games, the problems stated are of general nature, and we are convinced that our observations are applicable in the other domains as well.

The key takeaway is that having limited computational resources, it is probably better to stick with a simpler linear genome representation. Based on our research (which also supports intuition) they are more reliable to produce good solutions fast. 
However, with a large computational budget, we recommend applying a two-step approach.
After evolving vector-based solutions, transform them into equivalent trees and continue learning to take advantage of a more general model.

Another important observation is that self-improvement is potentially a better strategy than a predefined opponent when used as a goal of evolution.
Definitely, there is no point in learning against a weak opponent.
Learning against a strong opponent may be profitable but does not guarantee good performance in a broader context (e.g., tournament). 
And although progressive learning may also stagnate into some niche meta, it still seems to be more flexible in this aspect.

For future work, we plan to investigate a generalized bootstrapping-like scheme, that would switch between the representations automatically, as soon as the evolution progress drops below a certain threshold.
Separately, we would like to apply our approach to other tasks as well as evaluate different models, e.g., additional expert-knowledge features or trees with more operators.
When it comes to the different evolution schemes, some kind of ensemblement of both in-population evaluation and an external goal would be interesting. 
Also, we can consider the extension of using a portfolio of agents as the opponents, with some dynamic additions and removals, based on the win rates against the particular opponents.

\bibliographystyle{IEEEtran} 
\bibliography{bibliography} 

\end{document}